\documentclass[letterpaper, 10 pt, conference]{ieeeconf}
\IEEEoverridecommandlockouts                             
\usepackage{multirow}
\usepackage{cuted}
\usepackage{cite}

\usepackage{amsmath,amssymb,amsfonts}
\usepackage{algorithmic}
\usepackage{graphicx}
\usepackage{textcomp}
\usepackage{xcolor}
\usepackage{multirow}
\usepackage{diagbox}
\usepackage{graphicx}
\usepackage{makecell}
\usepackage{caption}
\usepackage{wrapfig,lipsum,booktabs}
\usepackage{tabularx}
\usepackage{float}
\usepackage{booktabs}
\usepackage{subcaption}
\usepackage[colorlinks = true, urlcolor  = blue]{hyperref}
\newcommand{\etal}{\textit{et al}.}
\newcommand{\citep}[1]{\cite{#1}}

\begin{document}

\title{\LARGE \bf Sim2Real Manipulation on Unknown Objects with Tactile-based Reinforcement Learning

}
\author{
Entong Su$^{1}$ , Chengzhe Jia$^{1,\dagger}$, Yuzhe Qin$^{1,\dagger}$, Wenxuan Zhou$^{2,\dagger}$, Annabella Macaluso$^{1,\dagger}$,\\Binghao Huang$^{3}$, Xiaolong Wang$^{1}$
\thanks{\textsuperscript{$1$} University of California San Diego, CA, USA}
\thanks{\textsuperscript{$2$} Carnegie Mellon University, PA, USA}
\thanks{\textsuperscript{$3$} University of Illinois Urbana-Champaign, IL, USA}
\thanks{$\dagger$ These authors contributed equally.}
}
\maketitle

\begin{abstract}
Using tactile sensors for manipulation remains one of the most challenging problems in robotics. At the heart of these challenges is generalization: How can we train a tactile-based policy that can manipulate unseen and diverse objects? In this paper, we propose to perform Reinforcement Learning with only visual tactile sensing inputs on diverse objects in a physical simulator. By training with diverse objects in simulation, it enables the policy to generalize to unseen objects. However, leveraging simulation introduces the Sim2Real transfer problem. To mitigate this problem, we study different tactile representations and evaluate how each affects real-robot manipulation results after transfer.  We conduct our experiments on diverse real-world objects and show significant improvements over baselines for the pivoting task. Our project page is available at \href{https://tactilerl.github.io/}{https://tactilerl.github.io/}.
\end{abstract}



\section{Introduction}
\begin{figure}[!htb]
\vspace{2mm}
    \begin{center}
    \begin{tabular}{c}
    \includegraphics[width=0.96\linewidth]{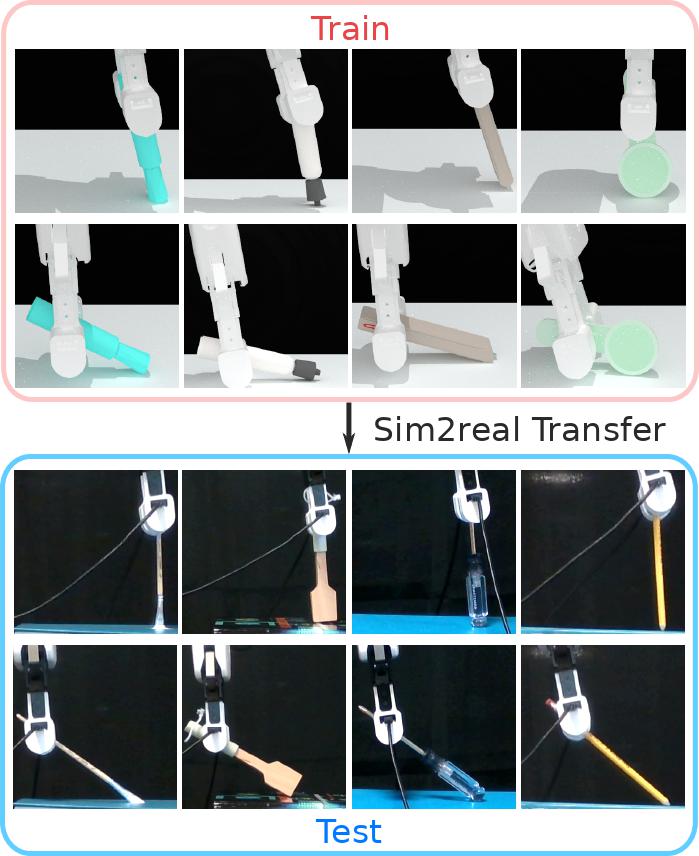} 
    \end{tabular}
    \end{center}
    \caption{
    We study the task of pivoting an object to a target angle with only tactile observations. Our tactile-based policy, trained with Reinforcement Learning purely in simulation, successfully transfers to the real robot without real-world data. The first row in each block visualizes the initial state of different episodes, while the second row demonstrates the final execution results.}
    \label{fig:cover}
\vspace{-5mm}
\end{figure}

When unlocking a door, we may reach into our bags for a key and re-orient it before inserting it into the lock to open the door; this series of actions are accomplished with a strong reliance on tactile cues. To enable robots to obtain similar skills, different tactile sensors~\citep{yuan2017gelsight,ward2018tactip,alspach2019softbubble,sferrazza2019design} have been designed and shown to be effective in capturing normal and shear forces for various manipulation tasks, especially when visual information is occluded. Following these works, the recent design of visual-tactile sensors~\citep{taylor2021gelslim30,gomes2020geltip, padmanabha2020omnitact} has further improved the resolution and sensitivity for capturing rich contact information and is more friendly to use with learning algorithms~\citep{villalonga2021tactile,zhong2022touching, zhang2020towards, higuera2023learning,ZHAO2023104321,kakani2021vision}. In this paper, we aim to explore the use of tactile sensors for the pivoting task as visualized in Figure~\ref{fig:cover}. The main challenge of the task is to understand the pose and geometry of the object and act accordingly based on tactile information at the fingertip of the robot gripper. 

One straightforward approach is to use tactile information to first estimate the object shape and pose, and directly use the estimated information as policy input~\citep{SHE_Cable,kim2023simultaneous,bauza2020tactile,gao2021objectfolder, Kupp2022tactile,pmlr-v205-sunil23a,wilson2023cable}. However, such estimations will likely be inaccurate given only partial observations,  which will introduce errors in the policy execution. Instead of explicit estimation of object geometry, we can leverage end-to-end Reinforcement Learning (RL), which takes the visual tactile observations as inputs, and directly outputs the action for more robust manipulation~\citep{Hansen2022Visuotactile,church2022tactile,bi2021zeroshot}. However, executing RL policies on real robots requires a significant number of interactions, which increases along with the complexity of the task. This prevents the policy from learning and operating with diverse objects. 

In this paper, we propose a system to train an RL policy with tactile inputs in simulation with diverse objects and then perform zero-shot Sim2Real transfer to the real robot, as shown in Figure~\ref{fig:cover}. By leveraging simulation, we largely increase the number and diversity of objects the robot interacts with, which leads to a more robust policy. However, there is still a large Sim2Real gap in how the tactile image is formed between the simulator and the real-world tactile sensor. Instead of performing perfect alignment between sim and real sensor imaging, we propose to study how to extract sufficient information from the sensor and reduce the domain gap under the context of RL inputs. Specifically, we ablate the tactile inputs by using: (i) the original tactile reading from the sensor; (ii) the difference of tactile information between the current reading and a canonical reading with no force applied; (iii) binary tactile information processed from the original tactile reading. The final version of observation provides less object information but turns out to reduce the Sim2Real gap effectively.

We experiment with different tactile representations on the tasks of pivoting~\cite{hou2018fast, hou2019reorienting}. 
Different from previous works that experiment with limited objects, we perform training and testing on a large number of objects, including 22 training objects in the simulation and 16 unseen objects for testing in the real world. By leveraging diverse training objects, the policy achieves better real-robot performance on a more abstract level of tactile representation compared to the full tactile information. To the best of our knowledge, our work is to conduct  Sim2Real transfer and achieve diverse object generalization for tactile-based manipulation. 

\section{Related Work}
\label{sec:citations}
\textbf{Vision-based Tactile Sensing.}
In recent years, vision-based tactile sensing has gained significant attention to enhance robot interaction with the environment. These sensors offer many benefits over force-based tactile sensors, such as increased spatial resolution and more detailed contact geometry. GelSight~\citep{yuan2017gelsight} is a pioneering work in this field, utilizing elastomeric material and a light-camera system to capture contact geometry. Following this, researchers have developed various methods to improve the design of such sensors, including OmniTact~\citep{padmanabha2020omnitact}, GelSlim~\citep{taylor2021gelslim30}, TacTip~\citep{ward2018tactip}, GelTip~\citep{gomes2020geltip}, and Digit~\cite{DIGIT_2020}.
To make use of the high-resolution data from these sensors, previous methods first built a state estimator from the tactile image and then integrated it with model-based control. For example, Oller~\etal~\citep{oller2022manipulation} uses Iterative Closest Points to estimate the object pose from contact points for down-stream manipulation task; Wilson~\etal~\citep{wilson2023cable} and She~\etal~\citep{SHE_Cable} use Principal Component Analysis to estimate the orientation of contact shape from GelSight reading for cable routing task. These methods encode the information from high-resolution images as low-dimensional states, which does not fully capitalize on the detailed contact geometry.
To better leverage the rich sensory data provided by vision-based tactile sensors, researchers employ end-to-end reinforcement learning for directly mapping tactile images to robot actions~\citep{Dong2021TactileRL, li2022see, van2015learning, xu2022towards}. However, the dynamics of tactile sensing are guided by complex contact mechanics, leading to a significant Sim2Real gap.
The work most closely related to ours is by Kim~\etal~\citep{kim2023simultaneous}, who tackled pivoting by estimating contact displacement in real-world conditions. Unlike their direct real-world application, our simulation-trained method utilizes diverse objects for broader applicability in the real world, handling unseen objects across categories by extracting key data from tactile images.

\textbf{Sim2Real Transfer for Tactile Sensing.}
Recently, several works have focused on bridging the gap between tactile simulation and real-world sensors. Yin~\etal\citep{yin2023rotating} proposed binarizing the Force Sensing Resistor (FSR) sensor signal to address in-hand rotating tasks, while Liang~\etal~\citep{liang2020hand} and Hebert~\etal\citep{hebert2011fusion} utilized binary contact modes to transfer policies on BioTac sensors for object pose tracking. These approaches effectively minimize the Sim2Real gap by converting both modalities into the binary domain. However, the process of binarizing tactile signals for vision-based tactile sensors with high spatial resolution remains unclear.
To obtain more realistic and dense tactile forces, recent work~\citep{narang2021interpreting, narang2021sim, ding2020sim, ma2019dense} combined `Finite Element Modeling' (FEM)~\citep{spyrakos1994finite} and learning-based methods to simulate tactile sensor deformations.
Although these simulators yield highly accurate tactile readings, their computational cost is prohibitively high, rendering them unsuitable for reinforcement learning (RL) training.
To enhance simulation efficiency, recent work~\citep{CycleGANGelSight, jianu2022reducing, zhao2023skill} train CycleGAN~\citep{zhu2017unpaired} on a self-collected dataset to convert tactile images across different domains. On the other side, Xu~\etal~\citep{xu2022efficient} introduced a penalty-based model featuring differentiable tactile simulation. While their method demonstrates promising Sim2Real transfer for robot manipulation, it has not been shown to generalize effectively to diverse and novel objects in the real world.
Rather than aligning sim and real sensor imaging with better simulation, our approach explores how various tactile image representations can reduce the domain gap.
Moreover, our method does not necessitate real-world training or data, thereby eliminating the need for additional data collection efforts.

\section{Sim2Real Transfer of Visual Tactile Readings}

\begin{figure}
    \vspace{2mm}
    \begin{center}
    \includegraphics[width=0.48\textwidth]{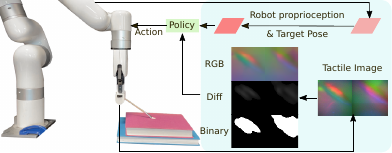} 
    \end{center}
   
    \caption{\textbf{Setup.} Two tactile sensors are mounted on the gripper's fingertips. Tactile readings are processed using our proposed approach and then serve as the input for training RL policies. }
    \label{fig:system-setup}
\vspace{-4mm}
\end{figure}

We propose a system designed for Sim2Real transfer of RL policies with tactile observations from the DIGIT tactile sensors (refer to Fig.~\ref{fig:system-setup}). Notably, the proposed system does not depend on any real-world data.
In this section, we will first explain how we generate visual-tactile data in the simulator to learn the pivoting task. Furthermore, to bridge the Sim2Real domain gap, we investigate three distinct representations to encode the tactile data. 


\subsection{Tactile Pattern Rendering in Simulation}
\label{sec:tac_sim}
In order to simulate the pattern of the DIGIT sensor, we extend the SAPIEN simulator\cite{Xiang_2020_SAPIEN} to accommodate visual tactile sensors, adhering to a pipeline similar to that in TACTO~\cite{Wang_2022}.
Three light sources are initialized, and the gel mesh is configured to align with the physical design of DIGIT. We update tactile images in real time using a linear mapping approach to convert resultant contact forces from the physics engine into deformation depth. This mapping enables object position adjustments based on applied normal forces. Then, we render RGB tactile images using Phong's model~\cite{gomes2021generation}. To speed up RL training, we replace TACTO's original OpenGL renderer with PyTorch3D. This modification enables direct GPU tensor rendering, eliminating the overhead for GPU-CPU data transfer and improving the  speed of image rendering.

\subsection{Tactile Image Representation}
\label{subsec:tactile_representation}
Performing RL with tactile input within a Sim2Real pipeline presents two main challenges: (i) the difficulty of simulating visual-tactile sensors accurately due to the Sim2Real gap and (ii) real-world DIGIT sensors' inconsistent behavior caused by manufacturing and gel variations. Therefore, instead of focusing on achieving perfect alignment in simulation, we investigate tactile image representations that can abstract the non-essential details in tactile images while preserving sufficient contact information that is crucial for robot decision-making.

Our study explores three tactile image representations, as illustrated in Figure~\ref{fig:system-setup}. The first representation is the original tactile RGB image directly acquired from the tactile simulation, as discussed in Section~\ref{sec:tac_sim} (denoted as \textbf{RGB}). 
Secondly, we propose a representation called \textbf{Diff}, which subtracts the current image from a force-free canonical image. This is done by calculating pixel-wise differences between the two RGB images and converting them to grayscale by averaging the RGB values.
Thirdly, we explore a binary version of the \textbf{Diff} image, which is denoted as \textbf{Binary}. This involves applying a predefined threshold ($\phi$) to distinguish between contacted and non-contacted pixels. 
We conduct a grid search to individually tailor the $\phi$ threshold for each digit to accommodate DIGIT's manufacturing variations and minimize noise. However, there is a trade-off between noise reduction and the possibility of missing certain contact information, as shown in Figure \ref{fig:pivot}. 

We augment the \textbf{Diff} and \textbf{Binary} tactile images by randomly scaling them between 0 and 1. 
We horizontally flip the tactile image from the right gripper to synchronize the data from the left and right grippers. This adjustment ensures that the angle information on the right side mirrors the left side, facilitating a uniform data representation. These tactile representations are used both in simulation and real-world experiments. 

\section{Learning Tactile Policies for Pivoting}
\label{policy_learning}

This section discusses how our proposed system uses tactile readings to train RL policies for the pivoting task~, as shown in Figure~\ref{fig:cover}. 

\textbf{Task Definition}: 
In the pivoting task, the robot needs to rotate the object to a target angle relative to the robot gripper pose. This operation exclusively depends on tactile sensing and joint proprioception, without any need for external sensors to estimate the object's pose.

\textbf{Observation Space:} 
The observation space contains tactile images, robot joint proprioceptive states, and task-related information (e.g., target angle). As mentioned in Section~\ref{subsec:tactile_representation}, we examine three different representations of tactile images, each with a resolution of $64 \times 64$.

\textbf{Action Space:}  
Since the focus of this work is on the robotic system rather than verifying the RL algorithm, we designed the action space to be as simple as possible to speed up RL training for this task. We restrict the end effector translation to the xz plane and restrict rotation to the y-axis.

The action space excludes the gripper width because it is initially set to grasp the object and remains constant throughout the manipulation process.

\begin{figure}
\vspace{2mm}
     \centering
     

      \begin{subfigure}[b]{0.47\textwidth}
    
         \includegraphics[width=\textwidth]{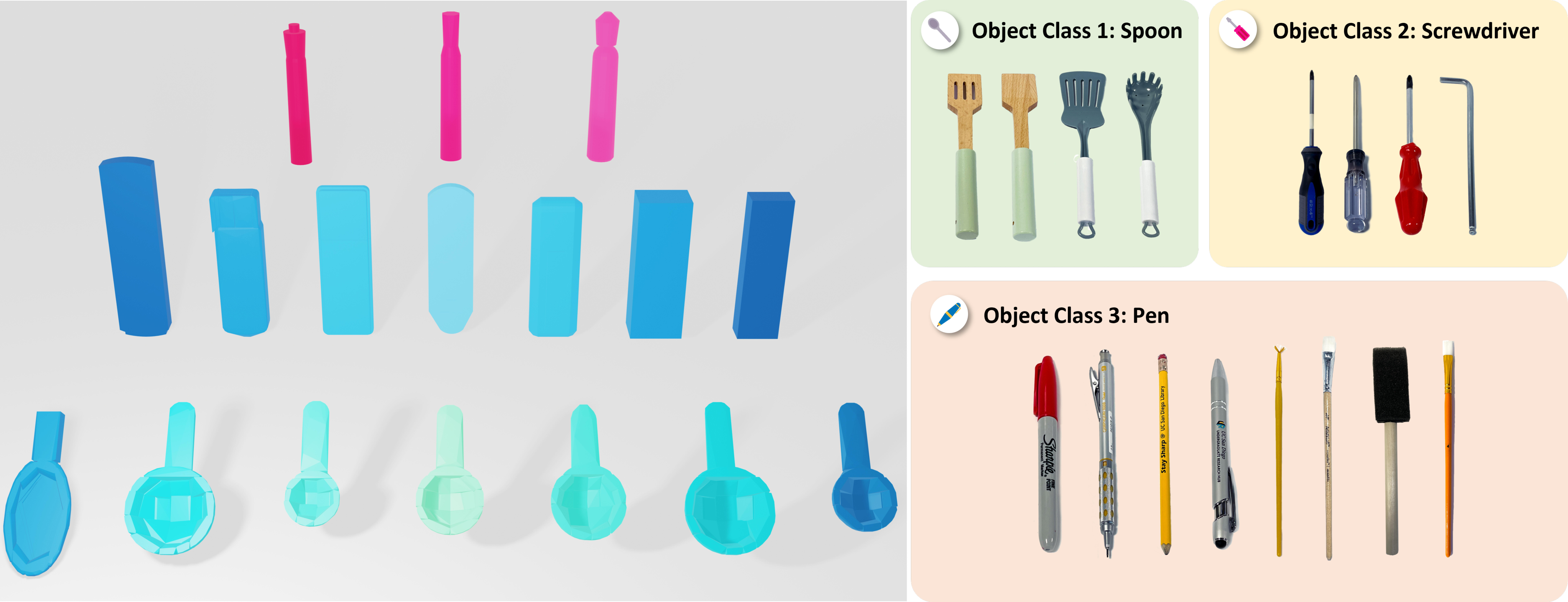}
         \caption{Simulation and real Objects}
         \label{fig:objects}
     \end{subfigure}

         
     

     \begin{subfigure}[b]{0.47\textwidth}
         \centering
         \includegraphics[width=\textwidth]{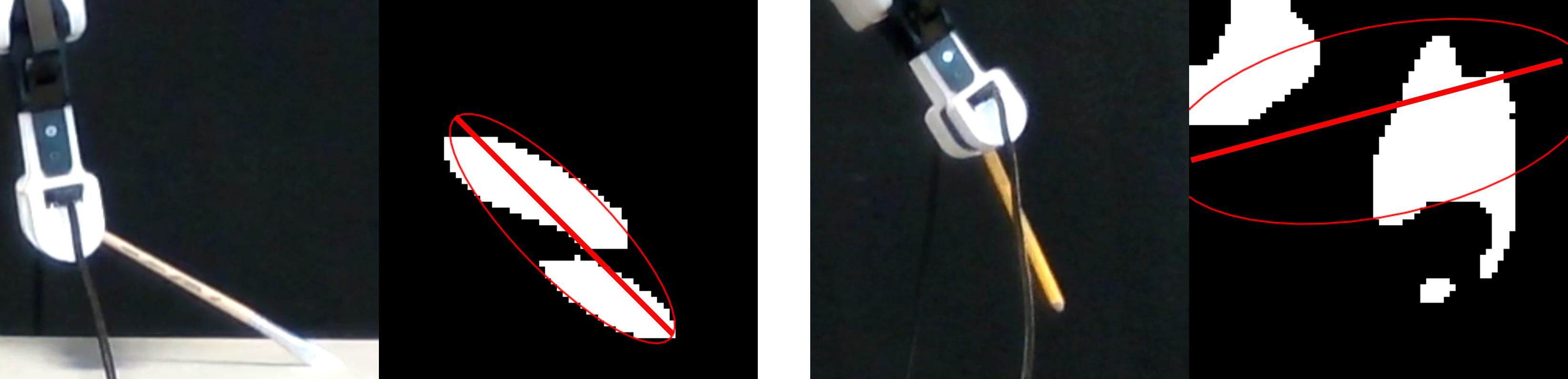}
         \caption{Angle estimation using PCA}
         \label{fig:pca}
     \end{subfigure}

     \begin{subfigure}[b]{0.47\textwidth}
         \centering
         \includegraphics[width=\textwidth]{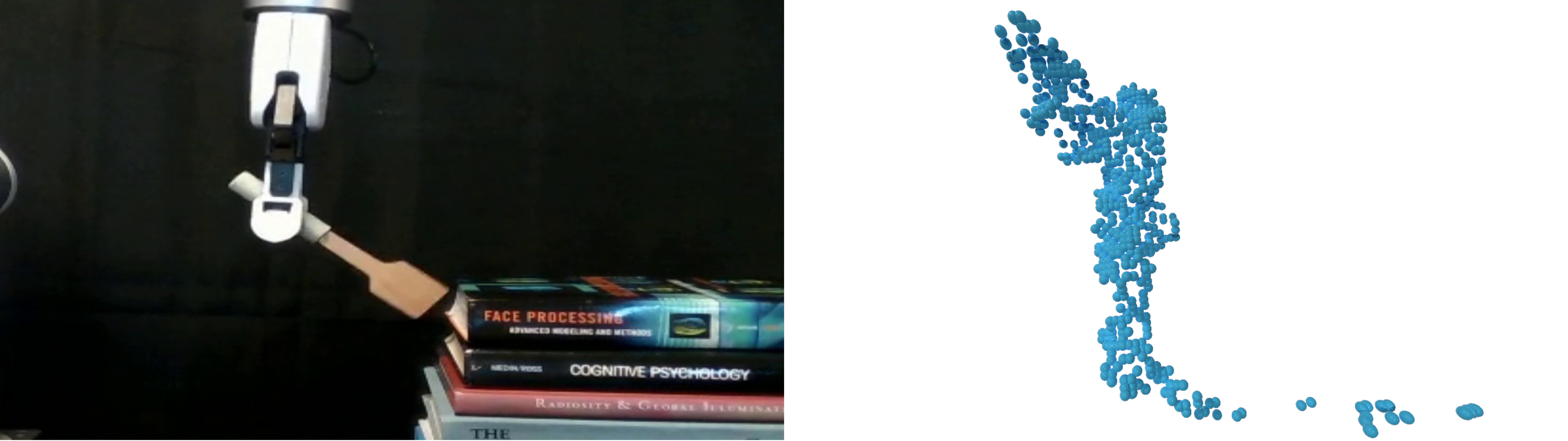}
         \caption{A failure case of the Point Cloud Policy}
         \label{fig:visual}
     \end{subfigure}

        \caption{\textbf{Object categories, Visualization of angle estimation using PCA and failure case of PC policy.}
We train with diverse objects in simulation (left image) and evaluate with a real robot (right image). In (b), we illustrate PCA angle estimation: a success (left) and a failure (right), with the red line indicating the estimated orientation. For (c), we present a failure case of the Point Cloud policy.
}
    \vspace{-3mm}  
        
\end{figure}

\textbf{Domain Randomization:} We use diverse objects from PartNet~\citep{mo2018partnet} and Breaking Bad~\citep{2022breaking}. The objects used for the simulation training and real evaluation are shown in Figure~\ref{fig:objects}. We randomize the height of the supporting surface from 0 to 20 cm relative to the robot base.
The object length ranges from 13 to 18 cm, and its initial pose varies between 165 and 195 degrees relative to the gripper. The target relative angle is randomized between 90 and 150 degrees. 

\textbf{Policy Training:} We use the Proximal Policy Optimization (PPO)~\cite{schulman2017proximal} for RL training. Two images are encoded using shared encoders and combined with the MLP feature of proprioception states. Both the actor and critic networks utilize the same feature.  We use five seeds for the training.  We use the default hyperparameter in \href{https://github.com/DLR-RM/stable-baselines3}{stable-baselines3} for the training.

\textbf{Reward Function:} The reward functions comprise four components: contact, distance-based, angle-based, and action penalties. The reward function is as follows:\\
$R=w_{contact}r_{contact}+w_{position}r_{position}+w_{angle}r_{angle}-w_{penalty}r_{penalty} $
\begin{itemize}
    \item Contact: The gripper earns a $+0.5$ reward ($r_{contact}$) upon object interaction, with $w_{contact}$ set to {0, 1, 2} depending on the tactile sensors' contact count. The goal is to sustain initial contact, reducing the chance of losing touch during rotation.
    \item Distance-based reward: 
Objects receive rewards for nearing the target position and penalties for distancing, using $w_{position}=10$ upon gripper contact. The position reward, $r_{position}$, varies between -1 and 1, computed as $r_{position} = 1 - (\frac{cur_{dist}}{init_{dist}})$, where $cur_{dist}$ and $init_{dist}$ represent the current and initial distances to the target position, respectively. This ensures objects maintain contact during rotation to reach the target angle.
    \item  Angle-based reward: This term is given by the difference between the current and target angles, following a structure similar to the distance-based reward.
    \item Action penalty: This term penalizes the magnitude of action output, defined as $r_{penalty} = ||a||^2$ with $w_{penalty} = 0.01$.
\end{itemize}

\textbf{Baselines:} We compared our proposed method with several baselines ranging from angle estimation, visual feedback, and expert demonstration. The baselines are categorized as follows:
\begin{enumerate}
    \item w/o Tactile: Only using proprioception observation.
    \item Oracle Angle: Trained with ground truth angles.
    \item Angle Estimator: 

We trained object-in-hand pose estimation using ConvNeXt\cite{liu2022convnet} as the backbone with tactile binary images. For real robot experiments, the estimated angle guides the Oracle Angle policy.
    \item PCA angle: PCA is employed to predict object orientation from tactile images, which is used on the Oracle angle policy.
    \item Point Cloud: We utilize point cloud data as observation for policy training, excluding the supporting table, and employ the PointNet architecture as the backbone.
    \item DAgger: The Tactile-Binary (Aug) policy guides the student policy, which doesn't rely on tactile information.
    \item  Tactile-RGB, ~Tactile-Depth, ~Tactile-Diff \textbf{(Ours)}: We utilize tactile readings without employing image augmentation during training.
    \item Tactile-RGB (Aug), \hspace{0.3pt}Tactile-Depth (Aug), \hspace{0.3pt}Tactile-Diff (Aug) \textbf{(Ours)}: We enhance our methods to incorporate image augmentation for tactile images, including the operations of random scale, erase. We introduce additional adjustments for RGB images on the brightness, contrast, and color hues.

\end{enumerate}

\textbf{Evaluation Metric}: We evaluate our policy based on two metrics: angle deviation and success rate. Angle deviation\footnote{We use the \href{https://www.amazon.com/gp/product/B0BJTWHXV3/ref=ppx_yo_dt_b_asin_title_o08_s00?ie=UTF8&th=1}{digital angle finder protractor} for the angle measurement in the real-world evaluation.} is the difference between the current and target rotated angles, expressed as a ratio. A task is considered successful when the angle deviation is under 15\%.
 Evaluation is based on the average success rate/error of these five seeds in the simulation. We evaluated 30 episodes in total in the real experiment.

\section{Sim2Real Transfer of Visual Tactile Readings}
\label{sec:method}
\begin{figure}[htbp]
\vspace{2mm}
    \begin{center}
    \begin{tabular}{c}
    \includegraphics[width=0.48\textwidth]{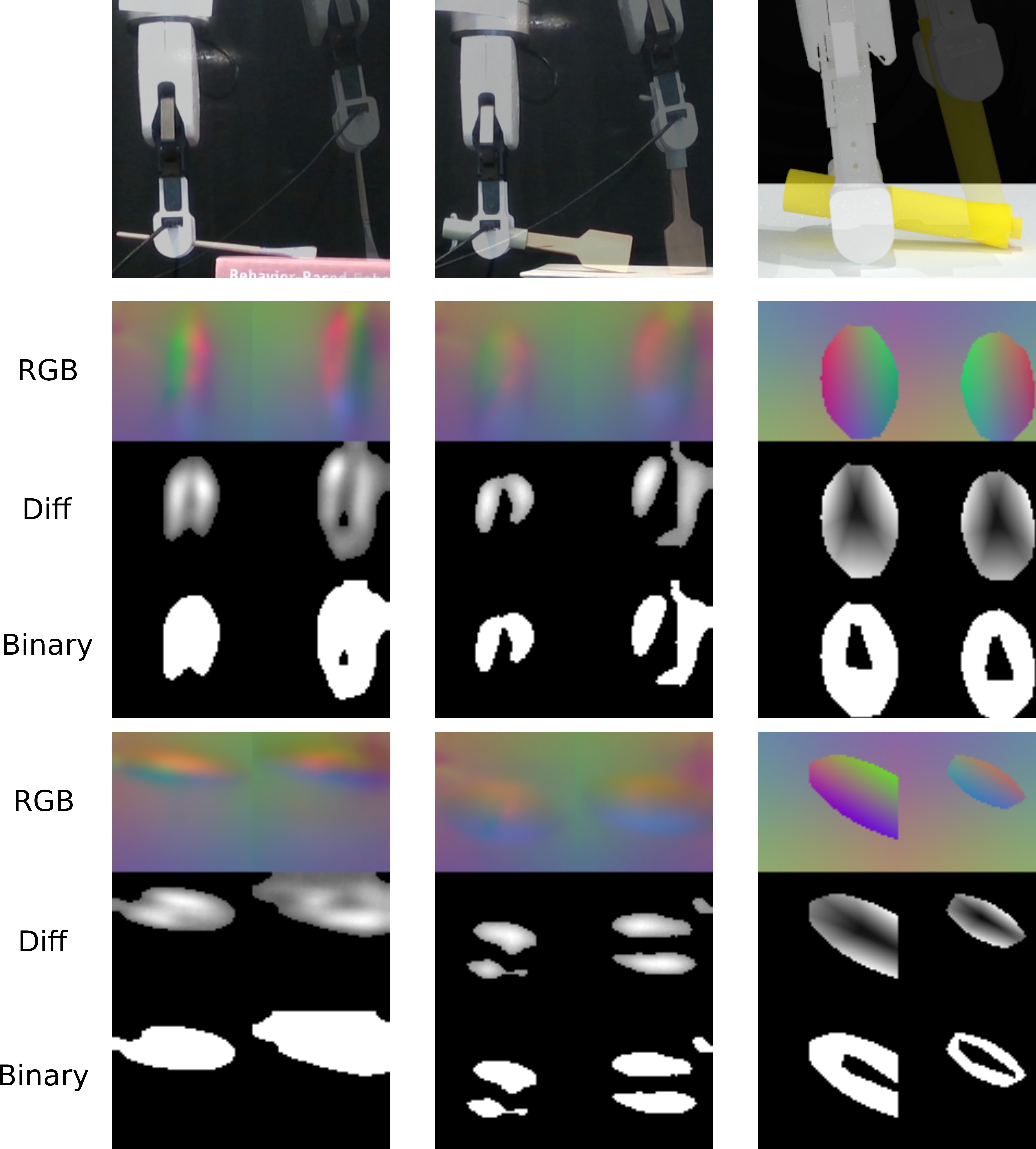} 
    \end{tabular}
    \end{center}
    
    \caption{\textbf{Real and Simulation Experiment for the pivoting.} We evaluated our pivoting task policy using tactile inputs in RGB, difference, and binary formats. The first two columns display the task's initial and final states on a real robot and in simulation. Rows two and three present tactile images from the start and end frames captured by both grippers. The last two rows sequentially showcase the RGB (RGB), difference (Diff), and binary (Binary) tactile images. }
   
    \label{fig:pivot}
     \vspace{-5mm}

\end{figure}

We conduct Sim2Real transfer experiments on the real robot to evaluate the effectiveness of our policy, as shown in Figure \ref{fig:pivot}. We directly transfer our policy to the real robot without any fine-tuning. In this section, we mainly investigate three aspects: 1) The necessity of tactile sensing, 2) The overall performance by using different tactile representations, and 3) The generalization of multi-category tactile policies.


\label{subsec:ablation-sim}

\begin{table*}[t]
\begin{center}
\vspace{2mm}
\scalebox{1.0}{
\centering
\begin{tabular}{ccccc} 
\toprule
 Environment &  \multicolumn{2}{c}{\textbf{Simulation}}  &  \multicolumn{2}{c}{\textbf{Real-World}} \\
Method & Deviation $\downarrow$ & Success $\uparrow $  & Deviation $\downarrow$ & Success $\uparrow $\\
\midrule

 w/o Tactile  & $34.84\% \pm$35.38\% &$0.32 \pm0.23$ & $23.55\%\pm8.00\%$ & $0.33\pm0.09$   \\

\textcolor{black}{DAgger}  &$17.14\%\pm3.04\%$  & $0.66\pm0.05$& $18.05\%\pm10.04\%$& $0.50\pm0.09$ \\

 Angle Estimator  & --  &  -- & $21.66\%\pm10.98\%$ & $0.60\pm0.08$ \\

 Point Cloud  &$16.2\%\pm2.14\%$  & $0.62\pm0.010$& $19.32\%\pm2.14\%$ &$0.50\pm0.11$ \\

 PCA Angle & ---- & --- & $30.19\%\pm14.32\%$ & $0.23\pm0.14$\\
 \textit{Oracle Angle} & $\mathit{9.22\% \pm1.43\%}$ & $\mathit{0.96\pm0.02}$ & --- &  ---\\

\midrule

  Tactile - RGB & $14.34\% \pm2.22\%$ & $0.64 \pm0.10$ & $18.03\%\pm7.42\%$ & $0.50\pm0.14$ \\
 Tactile - Diff & $15.67\% \pm3.33\%$ & $0.67 \pm0.10$ & $16.04\%\pm4.00\%$ & $0.60\pm0.04$ \\
 Tactile - Binary  & $15.31\% \pm 1.21\%$ &$0.65 \pm0.05$ & $12.25\%\pm3.70\%$ & $0.80\pm0.04$  \\
\midrule

\textcolor{black}{Tactile - RGB(Aug)}  &$13.19\%\pm2.77\%$  & $\mathbf{0.75\pm0.05}$& $11.56\%\pm6.44\%$& $0.76\pm0.06$ \\

\textcolor{black}{Tactile - Diff(Aug)}  &$14.35\%\pm2.49\%$  & $0.67\pm0.03$& $14.51\%\pm3.78\%$& $0.66\pm0.09$ \\

\textcolor{black}{Tactile - Binary(Aug)}  &$\mathbf{12.98\%\pm2.47\%}$  & $0.69\pm0.07$& $\mathbf{11.15\%\pm3.34\%}$& $\mathbf{0.80\pm0.02}$ \\

\bottomrule
\end{tabular}}
\caption{\textbf{Comparison of different observation modalities and tactile representations.} 
We assess angle deviation and success rate across simulated and real environments for policies trained with various inputs: three tactile image types (RGB, Diff., Binary), visual feedback (Point Cloud), expert demonstration (DAgger), angle metrics (Oracle Angle, PCA Angle, Angle Estimator), and scenarios without tactile feedback (w/o Tactile). The notation \textbf{(Aug)} signifies image augmentation for tactile representations, with Oracle Angle serving as an upper bound in simulation only.}
\label{table:sim-real-abation}
\end{center}
\vspace{-5mm}

\end{table*}

\begin{table*}[t]
\begin{center}
\scalebox{1.0}{

\begin{tabular}{c c c c c c c } 
 \\ 
\toprule
 Test Object type &\multicolumn{2}{c}{\textbf{Method}}   & 
 \multicolumn{2}{c}{\textbf{Simulation}} &  \multicolumn{2}{c}{\textbf{Real-World}}  

 \\

 & Observations & Objects & Deviation$\downarrow$ & Success $\uparrow$ & Deviation$\downarrow$ & Success $\uparrow$
 \\
\midrule
\multirow{2}{*}{Single Category}&w/o Tactile & Single Category & $24.45\%\pm14.37\%$ & $0.52\pm0.37$  & $26.68\%\pm8.01\%$ & $0.36\pm0.09$   \\

&Tactile - Binary(Aug) (Ours) & Single Category  & $\mathbf{8.23\%\pm 1.14\% }$  & $\mathbf{0.91\pm0.02}$ & $\mathbf{14.03\%\pm 4.60\%}$   
& $\mathbf{0.54\pm0.07}$  \\





\midrule

\multirow{6}{*}{Multi Category}&w/o Tactile & Single Category  & $72.12\%\pm11.58\%$ & $0.30\pm0.05$ &  $23.55\%\pm8.0\%$ & $0.33\pm0.16$  \\

&Tactile - Binary(Aug) (Ours) & Single Category  & $46.71\%\pm13.99\%$ & $0.42\pm0.13$ & $21.72\%\pm6\%$ & $0.53\pm0.19$  \\

\cmidrule{2-7}

&w/o Tactile & Multi Category & $34.84\%\pm35.38\%$& $0.32\pm0.23$ &$23.55\%\pm8.00\%$ & $0.33\pm0.09$     \\

&Tactile - Binary(Aug) (Ours) & Multi Category &  $\mathbf{12.98\%\pm2.47\%}$ &$\mathbf{0.69\pm 0.07}$  & $\mathbf{11.15\%\pm3.34\%}$ & $\mathbf{0.80\pm0.08}$ \\

\hline
\end{tabular}}

\end{center}

\caption{\textbf{Effect of multi objects training in simulation and real-world.} We contrast evaluation results for single and multi-object category training in pivoting tasks based on Tactile-Binary(Aug) and w/o Tactile policy.In the "Method" section, the term "object" refers to the items utilized during the training of the policy.}
\label{table:sim-SingleVSMulti}
\vspace{-3mm}
\end{table*}

\subsection{Necessity of Tactile Sensing}

\begin{figure}
     \vspace{1mm}
     \centering
     \includegraphics[width=0.48\textwidth]{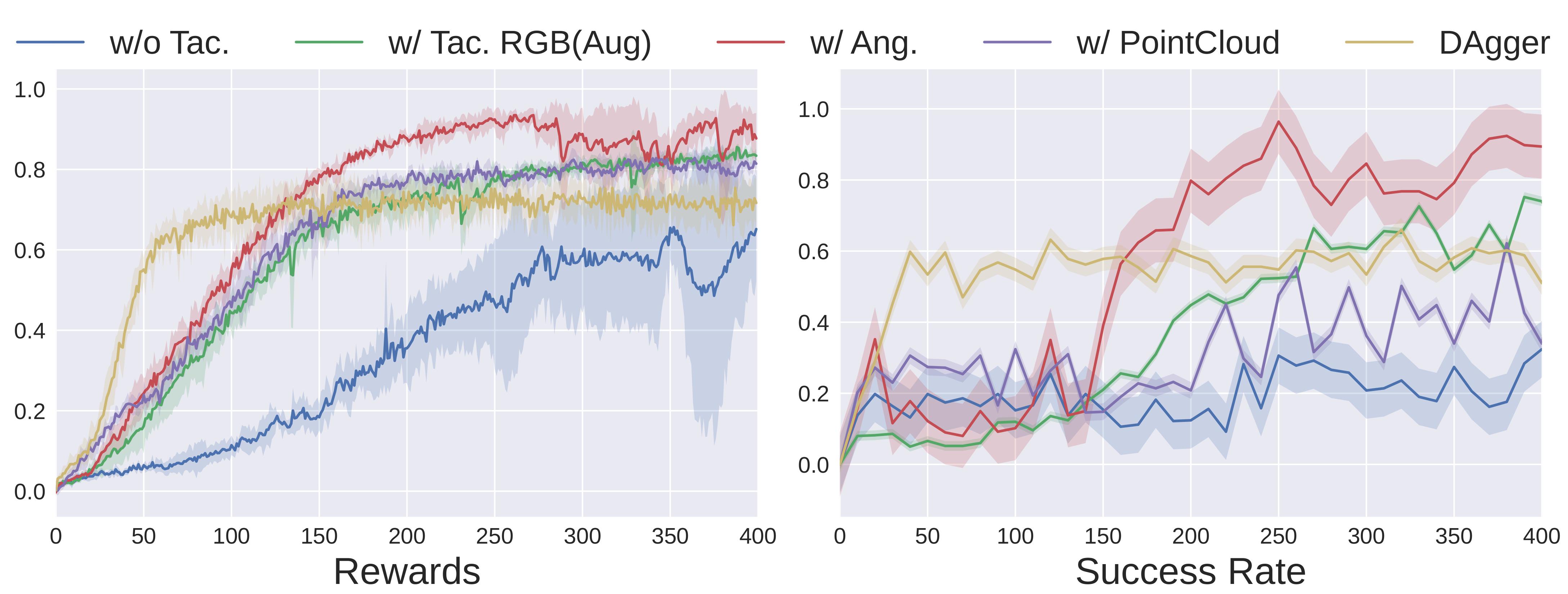}

        \caption{\textbf{Training curves.} We report the training curves for each task with two metrics: reward and success rate in the simulation.  Given that the tactile policies exhibit similar reward and success rate trends, we present the results for Tac. RGB (Aug) for simplicity. Note that \textit{Oracle Angle} achieves the best performance because it uses the ground truth object pose, which serves as an upper bound in the simulation experiment.}
        \label{fig:training-curves}
    \vspace{-5mm}
\end{figure}

To assess the importance of tactile sensing, we compare our methods with various baseline approaches described in Section \ref{policy_learning}. These baselines include observations such as angle estimation, visual feedback, and expert demonstration. Figure \ref{fig:training-curves} and  Table~\ref{table:sim-real-abation} summarize the training curves and evaluation results. 

\textbf{Comparison to w/o tactile}: We established a baseline for the tasks without tactile feedback. Compared with other methods, this baseline exhibited poorer performance, with lower success rates and larger angle deviations (first row in Table~\ref{table:sim-real-abation}). Figure \ref{fig:training-curves} shows that the policy without tactile feedback learns slowly and shows a higher variance in reward and success rate. This highlights the importance of visual feedback, state information, and tactile feedback for the pivoting.

\textbf{Comparison Against angle estimation methods}: As shown in Table~\ref{table:sim-real-abation}, the Oracle Angle policy achieves a 0.96 success rate in simulation, but this policy encounters the challenge of Sim2Real transfer. Using PCA and ConvNeXt for angle estimation achieves lower real-world success rates of 0.23 and 0.60, while Tactile-RGB(Aug) and Tactile-Binary(Aug) policies achieve approximately 0.80. Figure~\ref{fig:pca} illustrates the failure case of the PCA method. This gap indicates real-world factors, like noise, impacting angle estimation precision and task success.

\textbf{Comparison to Point Cloud policy}:  We also compared our policy with the policy solely relying on visual feedback. As shown in Table~\ref{table:sim-real-abation}, the Point Cloud policy's success rate drops from 0.62 in simulation to 0.50 in the real world. Figure \ref{fig:visual} shows the failure case of this policy. The main reason for these failures is the significant differences in size and shape between real-world objects and those in the simulation.

\textbf{Comparison to DAgger}: 
We used the tactile-binary policy, with a simulation success rate of 0.69, as the expert policy to guide the student policy without tactile information. The student policy achieved a similar success rate to the teacher policy in the simulation, but its real-world performance decreased to 0.50. This highlights the crucial role of tactile sensing in learning and decision-making during manipulation.

\begin{table*}
\begin{center}
\vspace{2mm}
\scalebox{1.0}{
\begin{tabular}{ccccc} 
\toprule
\textbf{Method} & \multicolumn{2}{c}{\textbf{Solid Table} }&\multicolumn{2}{c}{\textbf{Soft Table} } \\ 
  & Deviation $\downarrow$ & Success $\uparrow $  & Deviation $\downarrow$  & Success $\uparrow $ \\
\midrule
w/o Tactile & $23.55\%\pm8.00\%$ & $0.33\pm0.09$ & $22.26\%\pm11.70\%$ & $0.21\pm0.14$\\
Tactile - Binary (Ours) & $\mathbf{12.25\%\pm3.7}\%$ & $\mathbf{0.80\pm0.04}$& $\mathbf{11.60\%\pm 3.83\%}$ & $\mathbf{0.76\pm0.09}$\\
\bottomrule

\end{tabular}}
\end{center}

\caption{\textbf{Generalization to unseen supporting surfaces in the real world.} We summarize the angle deviation ratio and Real-World Pivoting Experiment success rate on different types of surfaces.}
\label{table:pivot_table}
\vspace{-3mm}
\end{table*}

\textbf{Evaluation Variation}: 
The tactile policy outperforms in the real world compared to simulations, likely due to evaluation setup differences: 500 episodes in simulation vs. 30 in the real world. Simulations feature more varied objects and scenarios, which real-world tests may not fully represent. This success in tactile policies suggests that tactile feedback improves decision-making in pivoting tasks by capturing essential angle information.

In summary, our tactile-based methods outperformed all others with the highest real-world success rate and the lowest deviation in the real-world evaluation. This result demonstrates the crucial role of tactile sensing within our system.

\subsection{Effect of Tactile Representations}
In this section, we explore the impact of policy training by utilizing different tactile representations(RGB, Binary, and Difference) and image augmentation on Sim2Real transfer. The evaluation results are summarized in Table~\ref{table:sim-real-abation}.

\textbf{Effect of different Tactile representations}: In our study, we analyzed various tactile representations without image augmentation, as detailed in Table~\ref{table:sim-real-abation} (lines 7 to 9). Despite achieving similar success rates (~0.65) and a 15\% angle deviation in simulation, real-world outcomes differed. The Tactile-RGB policy underperformed compared to others. Tactile-Binary showed consistent superiority in real-world tests, underscoring the sim2real challenges of RGB tactile data due to lighting, color, and pixel variations.


\textbf{Effect of Image augementation}: In Section \ref{policy_learning}, we enhanced tactile representations with image augmentation, with outcomes detailed in Table~\ref{table:sim-real-abation} (lines 10 to 12). Image augmentation notably boosted the tactile-RGB policy's success rates to 0.75 in simulation and 0.76 in real-world settings. Both Tactile-Binary and Tactile-Diff policies experienced marginal gains in performance across simulations and real environments, demonstrating image augmentation's role in improving tactile images' Sim2Real transferability.

In summary, employing image augmentation can yield significant benefits for Sim2Real transfer, particularly when dealing with RGB images. This is because image augmentation helps policies focus more on contact patterns rather than pixel values during training. As a result, it enhances the performance of policies relying on these three representations, resulting in more favorable outcomes.

\subsection{ Multi-Category Tactile Policy Generalization}

In this section, we explore how multi-object training improves the generalizability of our tactile system when evaluating multiple objects and previously unseen supporting surfaces.

\textbf{Effect of Multi-Objects Training:} We compared policies trained on single or multiple object categories for the pivoting task, and the result is summarized in Table~\ref{table:sim-SingleVSMulti}. Both training datasets contain multiple instances of each object category. 
The real-world evaluation reveals a marked decrease in the performance of single-category policies, highlighting their struggle to adapt to the wide variety of object geometries encountered in real-world scenarios.  In contrast, the multi-category tactile policy demonstrates superior generalization capabilities with a success rate of 0.80. This result highlights the multi-category policy's adaptability to diverse and unknown objects.

\textbf{Generalization to unseen supporting surfaces:} We also evaluated how well the policies can adapt to varying surfaces with different friction and stiffness properties. We evaluated this by using Tactile-RGB, Tactile-Binary, and Tactile-Binary policies. Each policy evaluated ten trials to ensure robustness and consistency, with results summarized in Table~\ref{table:pivot_table}. Despite a performance drop in both tactile-based and non-tactile methods, the tactile-based policy maintained a success rate of 0.76 with a 12\% angle deviation. This highlights the tactile-based policy can effectively adapt to various environmental conditions.


\textbf{Failure Cases}: There are two primary factors contributing to the failures of the tactile-based policy. The first is unstable gripping, leading to incomplete patterns. The second is incomplete or unusual contact, resulting in poor tactile feedback.

In summary, the result above demonstrates that our tactile-based policies are generalized to multi-categories and perform well on the previously unseen supporting surface in the real evaluation.

\section{Conclusion}
 In our study, we trained a tactile-based reinforcement learning policy for the pivoting task and successfully conducted Sim2Real transfer. The results show that our method is generalized to various unknown objects and previously unknown surfaces. 
 We are committed to releasing the code for our simulated environment and the training pipeline.

\clearpage
\bibliographystyle{IEEEtran}
\bibliography{IEEEabrv, Bibliography}

\begin{thebibliography}{10}
\providecommand{\url}[1]{#1}
\csname url@samestyle\endcsname
\providecommand{\newblock}{\relax}
\providecommand{\bibinfo}[2]{#2}
\providecommand{\BIBentrySTDinterwordspacing}{\spaceskip=0pt\relax}
\providecommand{\BIBentryALTinterwordstretchfactor}{4}
\providecommand{\BIBentryALTinterwordspacing}{\spaceskip=\fontdimen2\font plus
\BIBentryALTinterwordstretchfactor\fontdimen3\font minus \fontdimen4\font\relax}
\providecommand{\BIBforeignlanguage}[2]{{%
\expandafter\ifx\csname l@#1\endcsname\relax
\typeout{** WARNING: IEEEtran.bst: No hyphenation pattern has been}%
\typeout{** loaded for the language `#1'. Using the pattern for}%
\typeout{** the default language instead.}%
\else
\language=\csname l@#1\endcsname
\fi
#2}}
\providecommand{\BIBdecl}{\relax}
\BIBdecl

\bibitem{yuan2017gelsight}
W.~Yuan, S.~Dong, and E.~H. Adelson, ``Gelsight: High-resolution robot tactile sensors for estimating geometry and force,'' \emph{Sensors}, vol.~17, no.~12, p. 2762, 2017.

\bibitem{ward2018tactip}
B.~Ward-Cherrier, N.~Pestell, L.~Cramphorn, B.~Winstone, M.~E. Giannaccini, J.~Rossiter, and N.~F. Lepora, ``The tactip family: Soft optical tactile sensors with 3d-printed biomimetic morphologies,'' \emph{Soft robotics}, vol.~5, no.~2, pp. 216--227, 2018.

\bibitem{alspach2019softbubble}
A.~Alspach, K.~Hashimoto, N.~Kuppuswamy, and R.~Tedrake, ``Soft-bubble: A highly compliant dense geometry tactile sensor for robot manipulation,'' 2019.

\bibitem{sferrazza2019design}
C.~Sferrazza and R.~D’Andrea, ``Design, motivation and evaluation of a full-resolution optical tactile sensor,'' \emph{Sensors}, vol.~19, no.~4, p. 928, 2019.

\bibitem{taylor2021gelslim30}
I.~Taylor, S.~Dong, and A.~Rodriguez, ``Gelslim3.0: High-resolution measurement of shape, force and slip in a compact tactile-sensing finger,'' 2021.

\bibitem{gomes2020geltip}
D.~F. Gomes, Z.~Lin, and S.~Luo, ``Geltip: A finger-shaped optical tactile sensor for robotic manipulation,'' in \emph{2020 IEEE/RSJ International Conference on Intelligent Robots and Systems (IROS)}.\hskip 1em plus 0.5em minus 0.4em\relax IEEE, 2020, pp. 9903--9909.

\bibitem{padmanabha2020omnitact}
A.~Padmanabha, F.~Ebert, S.~Tian, R.~Calandra, C.~Finn, and S.~Levine, ``Omnitact: A multi-directional high-resolution touch sensor,'' in \emph{2020 IEEE International Conference on Robotics and Automation (ICRA)}.\hskip 1em plus 0.5em minus 0.4em\relax IEEE, 2020, pp. 618--624.

\bibitem{villalonga2021tactile}
M.~B. Villalonga, A.~Rodriguez, B.~Lim, E.~Valls, and T.~Sechopoulos, ``Tactile object pose estimation from the first touch with geometric contact rendering,'' in \emph{Conference on Robot Learning}.\hskip 1em plus 0.5em minus 0.4em\relax PMLR, 2021, pp. 1015--1029.

\bibitem{zhong2022touching}
\BIBentryALTinterwordspacing
S.~Zhong, A.~Albini, O.~P. Jones, P.~Maiolino, and I.~Posner, ``Touching a ne{RF}: Leveraging neural radiance fields for tactile sensory data generation,'' in \emph{6th Annual Conference on Robot Learning}, 2022. [Online]. Available: \url{https://openreview.net/forum?id=No3mbanRlZJ}
\BIBentrySTDinterwordspacing

\bibitem{zhang2020towards}
Y.~Zhang, W.~Yuan, Z.~Kan, and M.~Y. Wang, ``Towards learning to detect and predict contact events on vision-based tactile sensors,'' in \emph{Conference on Robot Learning}.\hskip 1em plus 0.5em minus 0.4em\relax PMLR, 2020, pp. 1395--1404.

\bibitem{higuera2023learning}
C.~Higuera, B.~Boots, and M.~Mukadam, ``Learning to read braille: Bridging the tactile reality gap with diffusion models,'' 2023.

\bibitem{ZHAO2023104321}
\BIBentryALTinterwordspacing
Y.~Zhao, X.~Jing, K.~Qian, D.~F. Gomes, and S.~Luo, ``Skill generalization of tubular object manipulation with tactile sensing and sim2real learning,'' \emph{Robotics and Autonomous Systems}, vol. 160, p. 104321, 2023. [Online]. Available: \url{https://www.sciencedirect.com/science/article/pii/S092188902200210X}
\BIBentrySTDinterwordspacing

\bibitem{kakani2021vision}
V.~Kakani, X.~Cui, M.~Ma, and H.~Kim, ``Vision-based tactile sensor mechanism for the estimation of contact position and force distribution using deep learning,'' \emph{Sensors}, vol.~21, no.~5, p. 1920, 2021.

\bibitem{SHE_Cable}
\BIBentryALTinterwordspacing
Y.~She, S.~Wang, S.~Dong, N.~Sunil, A.~Rodriguez, and E.~Adelson, ``Cable manipulation with a tactile-reactive gripper,'' \emph{The International Journal of Robotics Research}, vol.~40, no. 12-14, pp. 1385--1401, 2021. [Online]. Available: \url{https://doi.org/10.1177/02783649211027233}
\BIBentrySTDinterwordspacing

\bibitem{kim2023simultaneous}
S.~Kim, D.~K. Jha, D.~Romeres, P.~Patre, and A.~Rodriguez, ``Simultaneous tactile estimation and control of extrinsic contact,'' 2023.

\bibitem{bauza2020tactile}
M.~Bauza, E.~Valls, B.~Lim, T.~Sechopoulos, and A.~Rodriguez, ``Tactile object pose estimation from the first touch with geometric contact rendering,'' 2020.

\bibitem{gao2021objectfolder}
R.~Gao, Y.-Y. Chang, S.~Mall, L.~Fei-Fei, and J.~Wu, ``Objectfolder: A dataset of objects with implicit visual, auditory, and tactile representations,'' \emph{arXiv preprint arXiv:2109.07991}, 2021.

\bibitem{Kupp2022tactile}
N.~Kuppuswamy, A.~Castro, C.~Phillips-Grafflin, A.~Alspach, and R.~Tedrake, ``Fast model-based contact patch and pose estimation for highly deformable dense-geometry tactile sensors,'' \emph{IEEE Robotics and Automation Letters}, vol.~5, no.~2, pp. 1811--1818, 2020.

\bibitem{pmlr-v205-sunil23a}
\BIBentryALTinterwordspacing
N.~Sunil, S.~Wang, Y.~She, E.~Adelson, and A.~R. Garcia, ``Visuotactile affordances for cloth manipulation with local control,'' in \emph{Proceedings of The 6th Conference on Robot Learning}, ser. Proceedings of Machine Learning Research, K.~Liu, D.~Kulic, and J.~Ichnowski, Eds., vol. 205.\hskip 1em plus 0.5em minus 0.4em\relax PMLR, 14--18 Dec 2023, pp. 1596--1606. [Online]. Available: \url{https://proceedings.mlr.press/v205/sunil23a.html}
\BIBentrySTDinterwordspacing

\bibitem{wilson2023cable}
A.~Wilson, H.~Jiang, W.~Lian, and W.~Yuan, ``Cable routing and assembly using tactile-driven motion primitives,'' 2023.

\bibitem{Hansen2022Visuotactile}
J.~Hansen, F.~Hogan, D.~Rivkin, D.~Meger, M.~Jenkin, and G.~Dudek, ``Visuotactile-rl: Learning multimodal manipulation policies with deep reinforcement learning,'' in \emph{2022 International Conference on Robotics and Automation (ICRA)}, 2022, pp. 8298--8304.

\bibitem{church2022tactile}
A.~Church, J.~Lloyd, N.~F. Lepora \emph{et~al.}, ``Tactile sim-to-real policy transfer via real-to-sim image translation,'' in \emph{Conference on Robot Learning}.\hskip 1em plus 0.5em minus 0.4em\relax PMLR, 2022, pp. 1645--1654.

\bibitem{bi2021zeroshot}
T.~Bi, C.~Sferrazza, and R.~D'Andrea, ``Zero-shot sim-to-real transfer of tactile control policies for aggressive swing-up manipulation,'' 2021.

\bibitem{hou2018fast}
Y.~Hou, Z.~Jia, and M.~T. Mason, ``Fast planning for 3d any-pose-reorienting using pivoting,'' in \emph{2018 IEEE International Conference on Robotics and Automation (ICRA)}.\hskip 1em plus 0.5em minus 0.4em\relax IEEE, 2018, pp. 1631--1638.

\bibitem{hou2019reorienting}
------, ``Reorienting objects in 3d space using pivoting,'' \emph{arXiv preprint arXiv:1912.02752}, 2019.

\bibitem{DIGIT_2020}
\BIBentryALTinterwordspacing
M.~Lambeta, P.-W. Chou, S.~Tian, B.~Yang, B.~Maloon, V.~R. Most, D.~Stroud, R.~Santos, A.~Byagowi, G.~Kammerer, D.~Jayaraman, and R.~Calandra, ``{DIGIT}: A novel design for a low-cost compact high-resolution tactile sensor with application to in-hand manipulation,'' \emph{{IEEE} Robotics and Automation Letters}, vol.~5, no.~3, pp. 3838--3845, jul 2020. [Online]. Available: \url{https://doi.org/10.1109%2Flra.2020.2977257}
\BIBentrySTDinterwordspacing

\bibitem{oller2022manipulation}
M.~Oller, M.~Planas, D.~Berenson, and N.~Fazeli, ``Manipulation via membranes: High-resolution and highly deformable tactile sensing and control,'' 2022.

\bibitem{Dong2021TactileRL}
S.~Dong, D.~K. Jha, D.~Romeres, S.~Kim, D.~Nikovski, and A.~Rodriguez, ``Tactile-rl for insertion: Generalization to objects of unknown geometry,'' in \emph{2021 IEEE International Conference on Robotics and Automation (ICRA)}, 2021, pp. 6437--6443.

\bibitem{li2022see}
H.~Li, Y.~Zhang, J.~Zhu, S.~Wang, M.~A. Lee, H.~Xu, E.~Adelson, L.~Fei-Fei, R.~Gao, and J.~Wu, ``See, hear, and feel: Smart sensory fusion for robotic manipulation,'' 2022.

\bibitem{van2015learning}
H.~Van~Hoof, T.~Hermans, G.~Neumann, and J.~Peters, ``Learning robot in-hand manipulation with tactile features,'' in \emph{2015 IEEE-RAS 15th International Conference on Humanoid Robots (Humanoids)}.\hskip 1em plus 0.5em minus 0.4em\relax IEEE, 2015, pp. 121--127.

\bibitem{xu2022towards}
H.~Xu, Y.~Luo, S.~Wang, T.~Darrell, and R.~Calandra, ``Towards learning to play piano with dexterous hands and touch,'' in \emph{2022 IEEE/RSJ International Conference on Intelligent Robots and Systems (IROS)}.\hskip 1em plus 0.5em minus 0.4em\relax IEEE, 2022, pp. 10\,410--10\,416.

\bibitem{yin2023rotating}
Z.-H. Yin, B.~Huang, Y.~Qin, Q.~Chen, and X.~Wang, ``Rotating without seeing: Towards in-hand dexterity through touch,'' \emph{arXiv preprint arXiv:2303.10880}, 2023.

\bibitem{liang2020hand}
J.~Liang, A.~Handa, K.~Van~Wyk, V.~Makoviychuk, O.~Kroemer, and D.~Fox, ``In-hand object pose tracking via contact feedback and gpu-accelerated robotic simulation,'' in \emph{2020 IEEE International Conference on Robotics and Automation (ICRA)}.\hskip 1em plus 0.5em minus 0.4em\relax IEEE, 2020, pp. 6203--6209.

\bibitem{hebert2011fusion}
P.~Hebert, N.~Hudson, J.~Ma, and J.~Burdick, ``Fusion of stereo vision, force-torque, and joint sensors for estimation of in-hand object location,'' in \emph{2011 IEEE International Conference on Robotics and Automation}.\hskip 1em plus 0.5em minus 0.4em\relax IEEE, 2011, pp. 5935--5941.

\bibitem{narang2021interpreting}
Y.~S. Narang, B.~Sundaralingam, K.~V. Wyk, A.~Mousavian, and D.~Fox, ``Interpreting and predicting tactile signals for the syntouch biotac,'' 2021.

\bibitem{narang2021sim}
Y.~Narang, B.~Sundaralingam, M.~Macklin, A.~Mousavian, and D.~Fox, ``Sim-to-real for robotic tactile sensing via physics-based simulation and learned latent projections,'' in \emph{2021 IEEE International Conference on Robotics and Automation (ICRA)}.\hskip 1em plus 0.5em minus 0.4em\relax IEEE, 2021, pp. 6444--6451.

\bibitem{ding2020sim}
Z.~Ding, N.~F. Lepora, and E.~Johns, ``Sim-to-real transfer for optical tactile sensing,'' in \emph{2020 IEEE International Conference on Robotics and Automation (ICRA)}.\hskip 1em plus 0.5em minus 0.4em\relax IEEE, 2020, pp. 1639--1645.

\bibitem{ma2019dense}
D.~Ma, E.~Donlon, S.~Dong, and A.~Rodriguez, ``Dense tactile force estimation using gelslim and inverse fem,'' in \emph{2019 International Conference on Robotics and Automation (ICRA)}.\hskip 1em plus 0.5em minus 0.4em\relax IEEE, 2019, pp. 5418--5424.

\bibitem{spyrakos1994finite}
C.~C. Spyrakos, \emph{Finite Element Modeling}.\hskip 1em plus 0.5em minus 0.4em\relax West Virginia Univ. Press Morgantown, WV, USA, 1994.

\bibitem{CycleGANGelSight}
W.~Chen, Y.~Xu, Z.~Chen, P.~Zeng, R.~Dang, R.~Chen, and J.~Xu, ``Bidirectional sim-to-real transfer for gelsight tactile sensors with cyclegan,'' \emph{IEEE Robotics and Automation Letters}, vol.~7, no.~3, pp. 6187--6194, 2022.

\bibitem{jianu2022reducing}
T.~Jianu, D.~F. Gomes, and S.~Luo, ``Reducing tactile sim2real domain gaps via deep texture generation networks,'' in \emph{2022 International Conference on Robotics and Automation (ICRA)}.\hskip 1em plus 0.5em minus 0.4em\relax IEEE, 2022, pp. 8305--8311.

\bibitem{zhao2023skill}
Y.~Zhao, X.~Jing, K.~Qian, D.~F. Gomes, and S.~Luo, ``Skill generalization of tubular object manipulation with tactile sensing and sim2real learning,'' \emph{Robotics and Autonomous Systems}, vol. 160, p. 104321, 2023.

\bibitem{zhu2017unpaired}
J.-Y. Zhu, T.~Park, P.~Isola, and A.~A. Efros, ``Unpaired image-to-image translation using cycle-consistent adversarial networks,'' in \emph{Proceedings of the IEEE international conference on computer vision}, 2017, pp. 2223--2232.

\bibitem{xu2022efficient}
\BIBentryALTinterwordspacing
J.~Xu, S.~Kim, T.~Chen, A.~R. Garcia, P.~Agrawal, W.~Matusik, and S.~Sueda, ``Efficient tactile simulation with differentiability for robotic manipulation,'' in \emph{6th Annual Conference on Robot Learning}, 2022. [Online]. Available: \url{https://openreview.net/forum?id=6BIffCl6gsM}
\BIBentrySTDinterwordspacing

\bibitem{Xiang_2020_SAPIEN}
F.~Xiang, Y.~Qin, K.~Mo, Y.~Xia, H.~Zhu, F.~Liu, M.~Liu, H.~Jiang, Y.~Yuan, H.~Wang, L.~Yi, A.~X. Chang, L.~J. Guibas, and H.~Su, ``{SAPIEN}: A simulated part-based interactive environment,'' in \emph{The IEEE Conference on Computer Vision and Pattern Recognition (CVPR)}, June 2020.

\bibitem{Wang_2022}
\BIBentryALTinterwordspacing
S.~Wang, M.~Lambeta, P.-W. Chou, and R.~Calandra, ``{TACTO}: A fast, flexible, and open-source simulator for high-resolution vision-based tactile sensors,'' \emph{{IEEE} Robotics and Automation Letters}, vol.~7, no.~2, pp. 3930--3937, apr 2022. [Online]. Available: \url{https://doi.org/10.1109%2Flra.2022.3146945}
\BIBentrySTDinterwordspacing

\bibitem{gomes2021generation}
D.~F. Gomes, P.~Paoletti, and S.~Luo, ``Generation of gelsight tactile images for sim2real learning,'' 2021.

\bibitem{mo2018partnet}
K.~Mo, S.~Zhu, A.~X. Chang, L.~Yi, S.~Tripathi, L.~J. Guibas, and H.~Su, ``Partnet: A large-scale benchmark for fine-grained and hierarchical part-level 3d object understanding,'' 2018.

\bibitem{2022breaking}
S.~Sellán, Y.-C. Chen, Z.~Wu, A.~Garg, and A.~Jacobson, ``Breaking bad: A dataset for geometric fracture and reassembly,'' 2022.

\bibitem{schulman2017proximal}
J.~Schulman, F.~Wolski, P.~Dhariwal, A.~Radford, and O.~Klimov, ``Proximal policy optimization algorithms,'' 2017.

\bibitem{liu2022convnet}
Z.~Liu, H.~Mao, C.-Y. Wu, C.~Feichtenhofer, T.~Darrell, and S.~Xie, ``A convnet for the 2020s,'' 2022.

\end{thebibliography}

\end{document}